\documentclass[runningheads]{llncs}
\usepackage{graphicx}
\usepackage{comment}
\usepackage{amsmath,amssymb} %
\usepackage{color}
\usepackage{subfig}
\usepackage{adjustbox}
\usepackage{hyperref}

\begin{document}
\pagestyle{headings}
\mainmatter
\def\ECCVSubNumber{2114}  %

\title{Lift, Splat, Shoot: Encoding Images from Arbitrary Camera Rigs by Implicitly Unprojecting to 3D} %

\newcommand{\JP}[1]{{\color{violet}{[Jonah: #1]}}}
\newcommand{\SF}[1]{{\color{magenta}{[Sanja: #1]}}}

\titlerunning{Lift, Splat, Shoot}
\author{Jonah Philion \hspace{1.5mm}
Sanja Fidler}
\authorrunning{J. Philion et al.}
\institute{NVIDIA \hspace{1.5mm} University of Toronto
\hspace{1.5mm} Vector Institute}
\maketitle

\begin{figure}[h!]
\begin{center}
    \includegraphics[width=\linewidth,trim=0 0 0 0,clip]{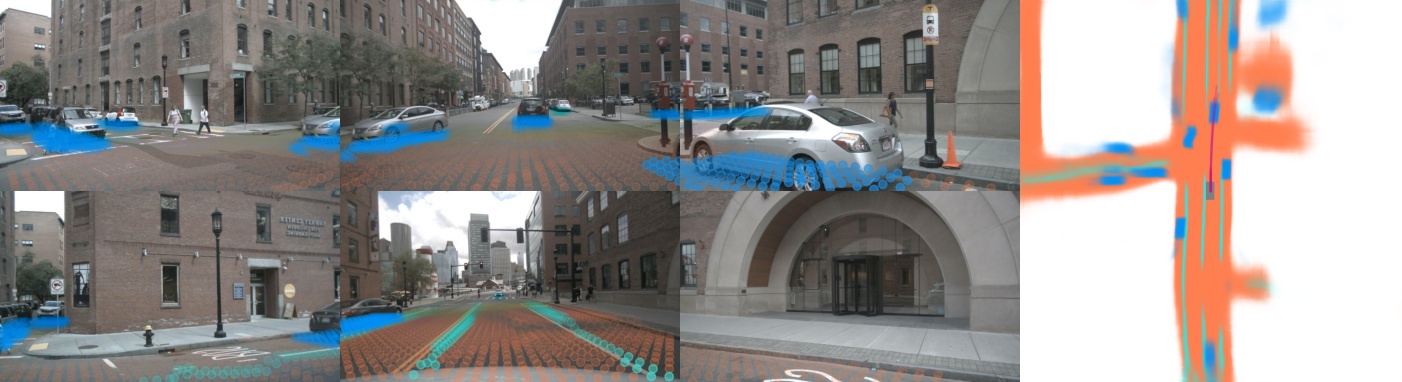}
\end{center}
    \caption{\footnotesize We propose a model that, given multi-view camera data (left), infers semantics directly in the bird's-eye-view (BEV) coordinate frame (right). We show vehicle segmentation (blue), drivable area (orange), and lane segmentation (green). These BEV predictions are then projected back onto input images (dots on the left).}%
\label{fig:hook}
\end{figure}

\begin{abstract}
The goal of perception for autonomous vehicles is to extract semantic representations from multiple sensors and fuse these representations into a single ``bird's-eye-view'' coordinate frame for consumption by motion planning. We propose a new end-to-end architecture that directly extracts a bird's-eye-view representation of a scene given image data from an arbitrary number of cameras. The core idea behind our approach is to ``lift'' each image individually into a frustum of features for each camera, then ``splat'' all frustums into a rasterized bird's-eye-view grid. By training on the entire camera rig, we provide evidence that our model is able to learn not only how to represent images but how to fuse predictions from all cameras into a single cohesive representation of the scene while being robust to calibration error. On standard bird's-eye-view tasks such as object segmentation and map segmentation, our model outperforms all baselines and prior work. In pursuit of the goal of learning dense representations for motion planning, we show that the representations inferred by our model enable interpretable end-to-end motion planning by ``shooting'' template trajectories into a bird's-eye-view cost map output by our network. We benchmark our approach against models that use oracle depth from lidar. Project page with code: {\color{magenta}{\href{https://nv-tlabs.github.io/lift-splat-shoot}{https://nv-tlabs.github.io/lift-splat-shoot}}}.
\end{abstract}

\section{Introduction}
\label{sec:intro}
Computer vision algorithms generally take as input an image and output either a prediction that is coordinate-frame agnostic -- such as in classification~\cite{mnist,imagenet,cifar,alexnet} -- or a prediction in the same coordinate frame as the input image -- such as in object detection, semantic segmentation, or panoptic segmentation~\cite{maskrcnn,segnet,panoptic,scnn}. 

\begin{figure}[t!]
\begin{center}
    \includegraphics[width=0.95\linewidth,trim=0 2 0 0,clip]{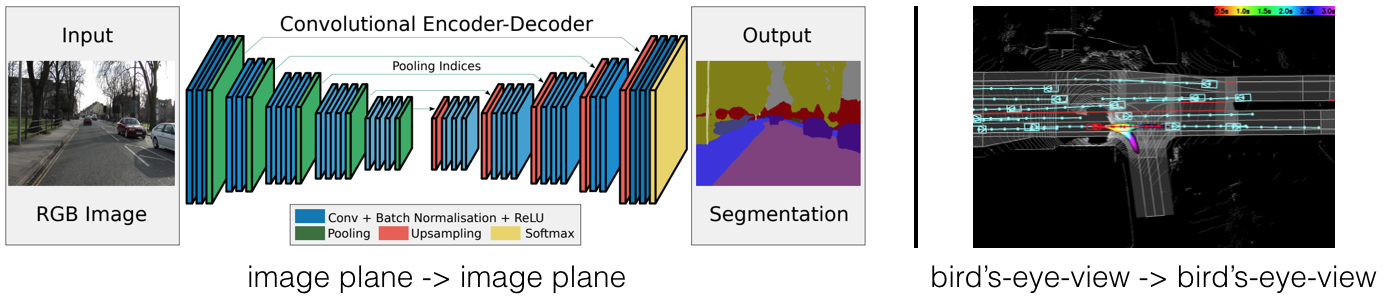}
\end{center}
    \caption{\footnotesize (left, from SegNet~\cite{segnet}) Traditionally, computer vision tasks such as semantic segmentation involve making predictions in the same coordinate frame as the input image. (right, from Neural Motion Planner~\cite{nmp}) In contrast, planning for self-driving generally operates in the bird's-eye-view frame. Our model directly makes predictions in a given bird's-eye-view frame for end-to-end planning from multi-view images.}
\label{fig:compare}
\end{figure}

This paradigm does not match the setting for perception in self-driving out-of-the-box. In self-driving, multiple sensors are given as input, each with a different coordinate frame, and perception models are ultimately tasked with producing predictions in a new coordinate frame -- the frame of the ego car -- for consumption by the downstream planner, as shown in Fig.~\ref{fig:compare}.

There are many simple, practical strategies for extending the single-image paradigm to the multi-view setting. For instance, for the problem of 3D object detection from $n$ cameras, one can apply a single-image detector to all input images individually, then rotate and translate each detection into the ego frame according to the intrinsics and extrinsics of the camera in which the object was detected. This extension of the single-view paradigm to the multi-view setting has three valuable symmetries baked into it:

\begin{enumerate}
    \item \textbf{Translation equivariance} -- If pixel coordinates within an image are all shifted, the output will shift by the same amount. Fully convolutional single-image object detectors roughly have this property and the multi-view extension inherits this property from them \cite{trans_invariance} \cite{goodfellow_text}.
    \item \textbf{Permutation invariance} -- the final output does not depend on a specific ordering of the $n$ cameras.
    \item \textbf{Ego-frame isometry equivariance} -- the same objects will be detected in a given image no matter where the camera that captured the image was located relative to the ego car. An equivalent way to state this property is that the definition of the ego-frame can be rotated/translated and the output will rotate/translate with it.
\end{enumerate}

The downside of the simple approach above is that using post-processed detections from the single-image detector prevents one from differentiating from predictions made in the ego frame all the way back to the sensor inputs. As a result, the model cannot learn in a data-driven way what the best way is to fuse information across cameras. It also means backpropagation cannot be used to automatically improve the perception system using feedback from the downstream planner.

We propose a model named ``Lift-Splat'' that preserves the 3 symmetries identified above by design while also being end-to-end differentiable. In Section~\ref{sec:method}, we explain how our model ``lifts'' images into 3D by generating a frustum-shaped point cloud of contextual features, ``splats'' all frustums onto a reference plane as is convenient for the downstream task of motion planning. In Section~\ref{subsec:explan}, we propose a method for ``shooting'' proposal trajectories into this reference plane for interpretable end-to-end motion planning. In Section~\ref{sec:impl}, we identify implementation details for training lift-splat models efficiently on full camera rigs. We present empirical evidence in Sec~\ref{sec:experiments} that our model learns an effective mechanism for fusing information from a distribution of possible inputs.

\section{Related Work}
\label{sec:related}

Our approach for learning cohesive representations from image data from multiple cameras builds on recent work in sensor fusion and monocular object detection. Large scale multi-modal datasets from Nutonomy \cite{nuscenes}, Lyft \cite{lyftlevel5}, Waymo \cite{waymo}, and Argo \cite{argo}, have recently made full representation learning of the entire $360^\circ$ scene local to the ego vehicle conditioned exclusively on camera input a possibility. We explore that possibility with our Lift-Splat architecture.

\subsection{Monocular Object Detection}
Monocular object detectors are defined by how they model the transformation from the image plane to a given 3-dimensional reference frame. A standard technique is to apply a mature 2D object detector in the image plane and then train a second network to regress 2D boxes into 3D boxes \cite{ssd16,fastsingle,mair,monogr}. The current state-of-the-art 3D object detector on the nuScenes benchmark \cite{mair} uses an architecture that trains a standard 2d detector to also predict depth using a loss that seeks to disentangle error due to incorrect depth from error due to incorrect bounding boxes. These approaches achieve great performance on 3D object detection benchmarks because detection in the image plane factors out the fundamental cloud of ambiguity that shrouds monocular depth prediction. 

An approach with recent empirical success is to separately train one network to do monocular depth prediction and another to do bird's-eye-view detection separately \cite{pseudolidar} \cite{pseudolidarpp}. These approaches go by the name of ``pseudolidar''. The intuitive reason for the empirical success of pseudolidar is that pseudolidar enables training a bird's-eye-view network that operates in the coordinate frame where the detections are ultimately evaluated and where, relative to the image plane, euclidean distance is more meaningful.

A third category of monocular object detectors uses 3-dimensional object primitives that acquire features based on their projection onto all available cameras. Mono3D \cite{mono3d} achieved state of the art monocular object detection on KITTI by generating 3-dimensional proposals on a ground plane that are scored by projecting onto available images. Orthographic Feature Transform \cite{oft} builds on Mono3D by projecting a fixed cube of voxels onto images to collect features and then training a second ``BEV'' CNN to detect in 3D conditioned on the features in the voxels. A potential performance bottleneck of these models that our model addresses is that a pixel contributes the same feature to every voxel independent of the depth of the object at that pixel.

\subsection{Inference in the Bird's-Eye-View Frame}
Models that use extrinsics and intrinsics in order to perform inference directly in the bird's-eye-view frame have received a large amount of interest recently. MonoLayout~\cite{krishna} performs bird's-eye-view inference from a single image and uses an adversarial loss to encourage the model to inpaint plausible hidden objects. In concurrent work, Pyramid Occupancy Networks~\cite{roddick} proposes a transformer architecture that converts image representations into bird's-eye-view representations. FISHING Net \cite{zoox} - also concurrent work - proposes a multi-view architecture that both segments objects in the current timestep and performs future prediction. We show that our model outperforms prior work empirically in Section~\ref{sec:experiments}. These architectures, as well as ours, use data structures similar to ``multi-plane'' images from the machine learning graphics community~\cite{splatnet,lighthouse,tucker2020singleview,neuralvolume}.

\begin{figure}[t!]
\centering
\includegraphics[height=3.4cm]{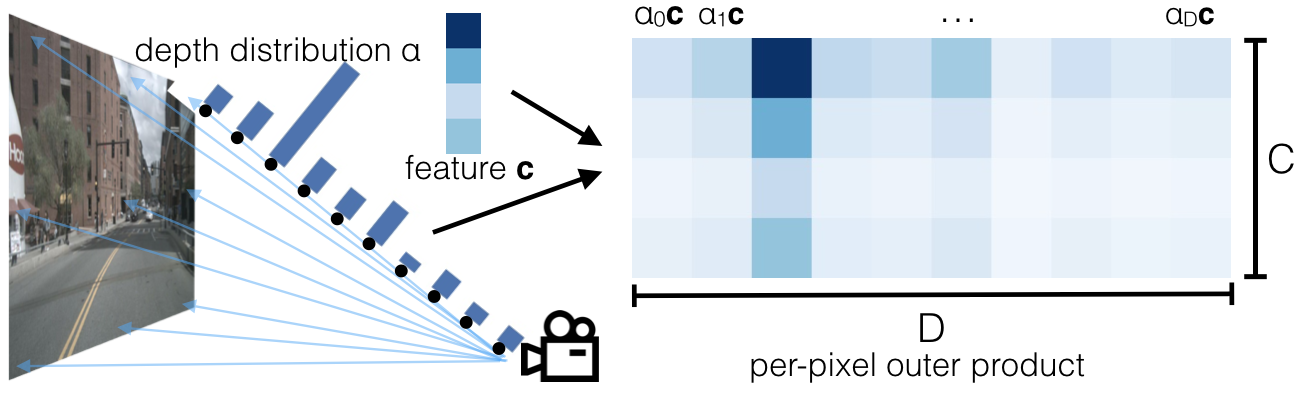}
\caption{\footnotesize We visualize the ``lift'' step of our model. For each pixel, we predict a categorical distribution over depth $\alpha \in \triangle^{D-1}$ (left) and a context vector $\mathbf{c} \in \mathbb{R}^C$ (top left). Features at each point along the ray are determined by the outer product of $\alpha$ and $\mathbf{c}$ (right).}
\label{fig:explain}
\end{figure}

\section{Method}
\label{sec:method}
 In this section, we present our approach for learning bird's-eye-view representations of scenes from image data captured by an arbitrary camera rig. We design our model such that it respects the symmetries identified in Section~\ref{sec:intro}.
 
 Formally, we are given $n$ images $\{\mathbf{X}_k \in \mathbb{R}^{3 \times H \times W}\}_n$ each with an extrinsic matrix $\mathbf{E}_k \in \mathbb{R}^{3 \times 4}$ and an intrinsic matrix $\mathbf{I}_k \in \mathbb{R}^{3 \times 3}$, and we seek to find a rasterized representation of the scene in the BEV coordinate frame $\mathbf{y} \in \mathbb{R}^{C \times X \times Y}$. The extrinsic and intrinsic matrices together define the mapping from reference coordinates $(x, y, z)$ to local pixel coordinates $(h, w, d)$ for each of the $n$ cameras. We do not require access to any depth sensor during training or testing.

 \subsection{Lift: Latent Depth Distribution}
 \label{subsec:lift}
  The first stage of our model operates on each image in the camera rig in isolation. The purpose of this stage is to ``lift'' each image from a local 2-dimensional coordinate system to a 3-dimensional frame that is shared across all cameras.

 The challenge of monocular sensor fusion is that we require depth to transform into reference frame coordinates but the ``depth'' associated to each pixel is inherently ambiguous. Our proposed solution is to generate representations at all possible depths for each pixel.
 
 Let $\mathbf{X} \in \mathbb{R}^{3 \times H \times W}$ be an image with extrinsics $\mathbf{E}$ and intrinsics $\mathbf{I}$, and let $p$ be a pixel in the image with image coordinates $(h, w)$. We associate $|D|$ points $\{ (h, w, d) \in \mathbb{R}^3 \mid d \in D \}$ to each pixel where $D$ is a set of discrete depths, for instance defined by $\{ d_0 + \Delta,..., d_0 + |D|\Delta \}$. Note that there are no learnable parameters in this transformation. We simply create a large point cloud for a given image of size $D \cdot H \cdot W$. This structure is equivalent to what the multi-view synthesis community~\cite{tucker2020singleview,lighthouse} has called a multi-plane image except in our case the features in each plane are abstract vectors instead of $(r,g,b,\alpha)$ values.
 
 The context vector for each point in the point cloud is parameterized to match a notion of attention and discrete depth inference. At pixel $p$, the network predicts a context $\mathbf{c}\in \mathbb{R}^C$ and a distribution over depth $\alpha \in \triangle^{|D| - 1}$ for every pixel. The feature $\mathbf{c}_d \in \mathbb{R}^C$ associated to point $p_d$ is then defined as the context vector for pixel $p$ scaled by $\alpha_d$:
 \begin{align}
    \mathbf{c}_d = \alpha_d \mathbf{c}.
 \end{align}

 Note that if our network were to predict a one-hot vector for $\alpha$, context at the point $p_d$ would be non-zero exclusively for a single depth $d^*$ as in pseudolidar \cite{pseudolidar}. If the network predicts a uniform distribution over depth, the network would predict the same representation for each point $p_d$ assigned to pixel $p$ independent of depth as in OFT \cite{oft}. Our network is therefore in theory capable of choosing between placing context from the image in a specific location of the bird's-eye-view representation versus spreading the context across the entire ray of space, for instance if the depth is ambiguous.

 In summary, ideally, we would like to generate a function $g_c : (x, y, z) \in \mathbb{R}^3 \rightarrow c \in \mathbb{R}^C$ for each image that can be queried at any spatial location and return a context vector. To take advantage of discrete convolutions, we choose to discretize space. For cameras, the volume of space visible to the camera corresponds to a frustum. A visual is provided in Figure~\ref{fig:explain}.

 \subsection{Splat: Pillar Pooling}
 \label{subsec:splat}

\begin{figure}[t]
\centering
\includegraphics[height=0.34\linewidth]{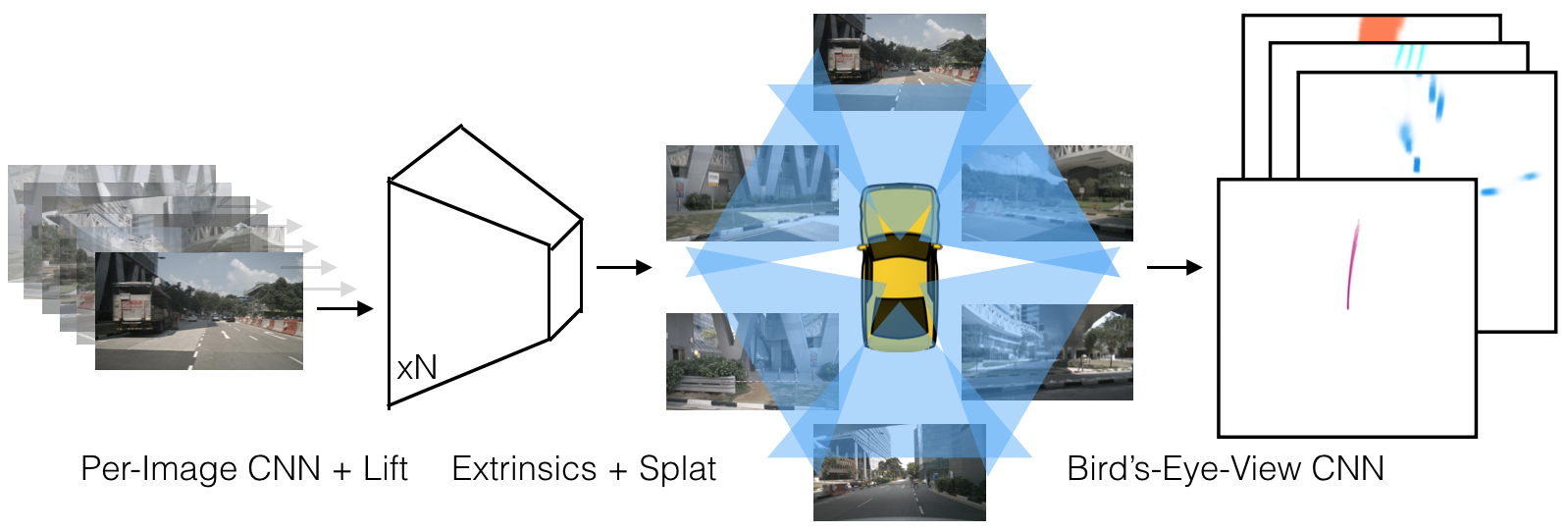}
\caption{\footnotesize \textbf{Lift-Splat-Shoot Outline} Our model takes as input $n$ images (left) and their corresponding extrinsic and intrinsic parameters. In the ``lift'' step, a frustum-shaped point cloud is generated for each individual image (center-left). The extrinsics and intrinsics are then used to splat each frustum onto the bird's-eye-view plane (center-right). Finally, a bird's-eye-view CNN processes the bird's-eye-view representation for BEV semantic segmentation or planning (right).}
\label{fig:splat}
\end{figure}

 We follow the pointpillars \cite{pointpillars} architecture to convert the large point cloud output by the ``lift'' step. ``Pillars'' are voxels with infinite height. We assign every point to its nearest pillar and perform sum pooling to create a $C \times H \times W$ tensor that can be processed by a standard CNN for bird's-eye-view inference. The overall lift-splat architecture is outlined in Figure~\ref{fig:splat}.

 Just as OFT \cite{oft} uses integral images to speed up their pooling step, we apply an analagous technique to speed up sum pooling. Efficiency is crucial for training our model given the size of the point clouds generated. Instead of padding each pillar then performing sum pooling, we avoid padding by using packing and leveraging a ``cumsum trick'' for sum pooling. This operation has an analytic gradient that can be calculated efficiently to speed up autograd as explained in subsection~\ref{subsec:customback}.

\subsection{Shoot: Motion Planning}
\label{subsec:explan}
    
Key aspect of our Lift-Splat model is that it enables end-to-end cost map learning for motion planning from camera-only input. At test time, planning using the inferred cost map can be achieved by ``shooting'' different trajectories, scoring their cost, then acting according to lowest cost trajectory \cite{pkl}. In Sec~\ref{subsec:planning}, we probe the ability of our model to enable end-to-end interpretable motion planning and compare its performance to lidar-based end-to-end neural motion planners.

We frame ``planning'' as predicting a distribution over $K$ template trajectories for the ego vehicle 
$$\mathcal{T} = \{\tau_i\}_K = \{\{ x_j, y_j, t_j \}_T\}_K$$
conditioned on sensor observations $p(\tau | o)$. Our approach is inspired by the recently proposed Neural Motion Planner (NMP)~\cite{nmp}, an architecture that conditions on point clouds and high-definition maps to generate a cost-volume that can be used to score proposed trajectories.

Instead of the hard-margin loss proposed in NMP, we frame planning as classification over a set of $K$ template trajectories. To leverage the cost-volume nature of the planning problem, we enforce the distribution over $K$ template trajectories to take the following form
\begin{align}
    p(\tau_i | o) = \frac{\exp\left(-\sum\limits_{x_i, y_i \in \tau_i} c_o(x_i, y_i) \right)}{\sum\limits_{\tau \in \mathcal{T}}\exp\left(-\sum\limits_{x_i, y_i \in \tau} c_o(x_i, y_i) \right)}
\end{align}
where $c_o(x, y)$ is defined by indexing into the cost map predicted given observations $o$ at location $x,y$ and can therefore be trained end-to-end from data by optimizing for the log probability of expert trajectories. For labels, given a ground-truth trajectory, we compute the nearest neighbor in L2 distance to the template trajectories $\mathcal{T}$ then train with the cross entropy loss. This definition of $p(\tau_i | o)$ enables us to learn an interpretable spatial cost function without defining a hard-margin loss as in NMP \cite{nmp}.

\begin{figure}[t!] 
\begin{minipage}{0.47\linewidth}
\begin{center}
    \includegraphics[width=1\linewidth,trim= 0 25 0 35,clip]{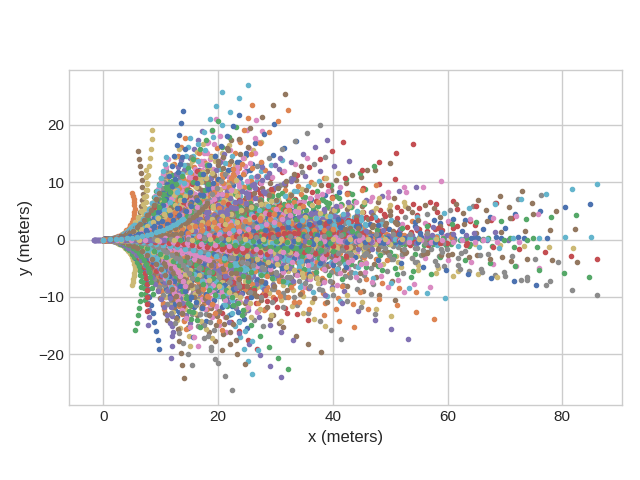}
\end{center}
\end{minipage}
\begin{minipage}{0.525\linewidth}
    \caption{\footnotesize We visualize the 1K trajectory templates that we ``shoot'' onto our cost map during training and testing. During training, the cost of each template trajectory is computed and interpreted as a 1K-dimensional Boltzman distribution over the templates. During testing, we choose the argmax of this distribution and act according to the chosen template.}
\label{fig:trajs}
\end{minipage}
\end{figure}

In practice, we determine the set of template trajectories by running K-Means on a large number of expert trajectories. The set of template trajectories used for ``shooting'' onto the cost map that we use in our experiments are visualized in Figure~\ref{fig:trajs}.

\section{Implementation}
\label{sec:impl}
\subsection{Architecture Details}
The neural architecture of our model is similar to OFT \cite{oft}. As in OFT, our model has two large network backbones. One of the backbones operates on each image individually in order to featurize the point cloud generated from each image. The other backbone operates on the point cloud once it is splatted into pillars in the reference frame. The two networks are joined by our lift-splat layer as defined in Section~\ref{sec:method} and visualize in Figure~\ref{fig:splat}.

For the network that operates on each image in isolation, we leverage layers from an EfficientNet-B0 \cite{efficientnet} pretrained on Imagenet \cite{imagenet} in all experiments for all models including baselines. EfficientNets are network architectures found by exhaustive architecture search in a resource limited regime with depth, width, and resolution scaled up proportionally. We find that they enable higher performance relative to ResNet-18/34/50 \cite{resnet} across all models with a minor inconvenience of requiring more optimization steps to converge.

For our bird's-eye-view network, we use a combination of ResNet blocks similar to PointPillars \cite{pointpillars}. Specifically, after a convolution with kernel 7 and stride 2 followed by batchnorm \cite{batchnorm} and ReLU \cite{relu}, we pass through the first 3 meta-layers of ResNet-18 to get 3 bird's-eye-view representations at different resolutions $x_1, x_2, x_3$. We then upsample $x_3$ by a scale factor of 4, concatenate with $x_1$, apply a resnet block, and finally upsample by 2 to return to the resolution of the original input bird's-eye-view pseudo image. We count 14.3M trainable parameters in our final network.

There are several hyper-parameters that determine the ``resolution'' of our model. First, there is the size of the input images $H \times W$. In all experiments below, we resize and crop input images to size $128 \times 352$ and adjust extrinsics and intrinsics accordingly. Another important hyperparameter of network is the size the resolution of the bird's-eye-view grid $X \times Y$. In our experiments, we set bins in both $x$ and $y$ from -50 meters to 50 meters with cells of size 0.5 meters $\times$ 0.5 meters. The resultant grid is therefore $200 \times 200$. Finally, there's the choice of $D$ that determines the resolution of depth predicted by the network. We restrict $D$ between 4.0 meters and 45.0 meters spaced by 1.0 meters. With these hyper-parameters and architectural design choices, the forward pass of the model runs at 35 hz on a Titan V GPU.

\subsection{Frustum Pooling Cumulative Sum Trick}
\label{subsec:customback}

Training efficiency is critical for learning from data from an entire sensor rig. We choose sum pooling across pillars in Section~\ref{sec:method} as opposed to max pooling because our ``cumulative sum'' trick saves us from excessive memory usage due to padding. The ``cumulative sum trick'' is the observation that sum pooling can be performed by sorting all points according to bin id, performing a cumulative sum over all features, then subtracting the cumulative sum values at the boundaries of the bin sections. Instead of relying on autograd to backprop through all three steps, the analytic gradient for the module as a whole can be derived, speeding up training by 2x. We call the layer ``Frustum Pooling'' because it handles converting the frustums produced by $n$ images into a fixed dimensional $C\times H \times W$ tensor independent of the number of cameras $n$. Code can be found on our \href{https://nv-tlabs.github.io/lift-splat-shoot/}{project page}.

\section{Experiments and Results}
\label{sec:experiments}

We use the nuScenes \cite{nuscenes} and Lyft Level 5 \cite{lyftlevel5} datasets to evaluate our approach. nuScenes is a large dataset of point cloud data and image data from 1k scenes, each of 20 seconds in length. The camera rig in both datasets is comprised of 6 cameras which roughly point in the forward, front-left, front-right, back-left, back-right, and back directions. In all datasets, there is a small overlap between the fields-of-view of the cameras. The extrinsic and intrinsic parameters of the cameras shift throughout both datasets. Since our model conditions on the camera calibration, it is able to handle these shifts.

We define two object-based segmentation tasks and two map-based tasks. For the object segmentation tasks, we obtain ground truth bird's-eye-view targets by projecting 3D bounding boxes into the bird's-eye-view plane. Car segmentation on nuScenes refers to all bounding boxes of class \verb|vehicle.car| and vehicle segmentation on nuScenes refers to all bounding boxes of meta-category \verb|vehicle|. Car segmentation on Lyft refers to all bounding boxes of class \verb|car| and vehicle segmentation on nuScenes refers to all bounding boxes with class $\in \{$ \verb|car, truck, other_vehicle, bus, bicycle| $\}$. For mapping, we use transform map layers from the nuScenes map into the ego frame using the provided 6 DOF localization and rasterize.

For all object segmentation tasks, we train with binary cross entropy with positive weight 1.0. For the lane segmentation, we set positive weight to 5.0 and for road segmentation we use positive weight 1.0 \cite{fastdraw}. In all cases, we train for 300k steps using Adam \cite{adam} with learning rate $1e-3$ and weight decay $1e-7$. We use the PyTorch framework \cite{pytorch}.

The Lyft dataset does not come with a canonical train/val split. We separate 48 of the Lyft scenes for validation to get a validation set of roughly the same size as nuScenes (6048 samples for Lyft, 6019 samples for nuScenes).

\subsection{Description of Baselines}
 Unlike vanilla CNNs, our model comes equipped with 3-dimensional structure at initialization. We show that this structure is crucial for good performance by comparing against a CNN composed of standard modules. We follow an architecture similar to MonoLayout~\cite{krishna} which also  trains a CNN to output bird's-eye-view labels from images only but does not leverage inductive bias in designing the architecture and trains on single cameras only. The architecture has an EfficientNet-B0 backbone that extracts features independently across all images. We concatenate the representations and perform bilinear interpolation to upsample into a $\mathbb{R}^{X \times Y}$ tensor as is output by our model. We design the network such that it has roughly the same number of parameters as our model. The weak performance of this baseline demonstrates how important it is to explicitly bake symmetry 3 from Sec~\ref{sec:intro} into the model in the multi-view setting.

 To show that our model is predicting a useful implicit depth, we compare against our model where the weights of the pretrained CNN are frozen as well as to OFT \cite{oft}. We outperform these baselines on all tasks, as shown in Tables~\ref{tab:obj} and~\ref{tab:map}. We also outperform concurrent work that benchmarks on the same segmentation tasks \cite{zoox} \cite{roddick}. As a result, the architecture is learning both an effective depth distribution as well as effective contextual representations for the downstream task.

\subsection{Segmentation}
\label{subsec:super}

We demonstrate that our Lift-Splat model is able to learn semantic 3D representations given supervision in the bird's-eye-view frame. Results on the object segmentation tasks are shown in Table~\ref{tab:obj}, while results on the map segmentation tasks are in Table~\ref{tab:map}. On all benchmarks, we outperform our baselines. We believe the extent of these gains in performance from implicitly unprojecting into 3D are substantial, especially for object segmentation. We also include reported IOU scores for two concurrent works \cite{zoox} \cite{roddick} although both of these papers use different definitions of the bird's-eye-view grid and a different validation split for the Lyft dataset so a true comparison is not yet possible.
    
    \begin{table}[t!]
    \begin{minipage}{0.47\linewidth}
    \begin{center}
     \begin{adjustbox}{width=\textwidth}
    \begin{tabular}{|l|c|c||c|c|}
    \cline{2-5}
    \multicolumn{1}{c|}{} & \multicolumn{2}{c||}{nuScenes} & \multicolumn{2}{c|}{Lyft}\\
    \cline{2-5}
    \multicolumn{1}{c|}{} & Car & Vehicles & Car & Vehicles\\
    \hline
    CNN & 22.78 & 24.25 & 30.71 & 31.91\\
    \hline
    Frozen Encoder & 25.51 & 26.83 & 35.28 & 32.42\\
    \hline
    OFT & 29.72 & 30.05 & 39.48 & 40.43\\
    \hline
    Lift-Splat (Us) & \textbf{32.06} & \textbf{32.07} & \textbf{43.09} & \textbf{44.64}\\
    \hline
    \hline
    PON$^*$ \cite{roddick} & 24.7 & - & - & -\\
    \hline
    FISHING$^*$ \cite{zoox} & - & 30.0 & - & 56.0\\
    \hline
    \end{tabular}
    \end{adjustbox}
    \end{center}
    \caption{\footnotesize Segment. IOU in BEV frame}
    \label{tab:obj}
    \end{minipage}
   \begin{minipage}{0.5\linewidth}
    \begin{center}
     \begin{adjustbox}{width=\textwidth}
    \begin{tabular}{|l|c|c|}
    \cline{2-3}
    \multicolumn{1}{c|}{} & Drivable Area & Lane Boundary \\
    \hline
    CNN & 68.96 & 16.51 \\
    \hline
    Frozen Encoder & 61.62 & 16.95 \\
    \hline
    OFT & 71.69 & 18.07  \\
    \hline
    Lift-Splat (Us) & \textbf{72.94} & \textbf{19.96} \\
    \hline
    \hline 
    PON$^*$ \cite{roddick} & 60.4 & - \\
    \hline
    \end{tabular}
     \end{adjustbox}
    \end{center}
    \caption{\footnotesize Map IOU in BEV frame}
    \label{tab:map}
     \end{minipage}
        \end{table}

\subsection{Robustness}
\label{subsec:robustness}
\begin{figure}[t]%
    \centering
    \subfloat[Test Time Extrinsic Noise]{{\includegraphics[width=5cm]{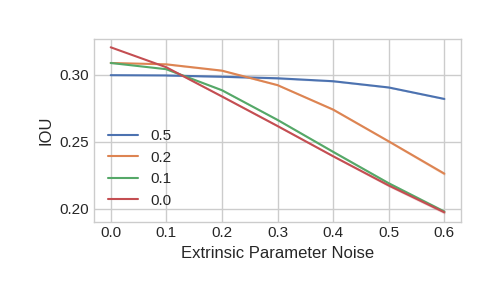} }}%
    \qquad
    \subfloat[Test Time Camera Dropout]{{\includegraphics[width=5cm]{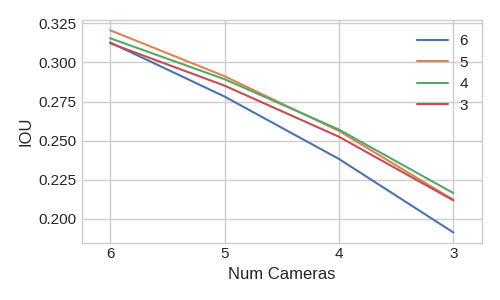} }}%
    \caption{\footnotesize We show that it is possible to train our network such that it is resilient to common sources of sensor error. On the left, we show that by training with a large amount of noise in the extrinsics (blue), the network becomes more robust to extrinsic noise at test time. On the right, we show that randomly dropping cameras from each batch during training (red) increases robustness to sensor dropout at test time.}%
    \label{fig:robust}%
\end{figure}

Because the bird's-eye-view CNN learns from data how to fuse information across cameras, we can train the model to be robust to simple noise models that occur in self-driving such as extrinsics being biased or cameras dying. In Figure~\ref{fig:robust}, we verify that by dropping cameras during training, our model handles dropped cameras at better at test time. In fact, the best performing model when all 6 cameras are present is the model that is trained with 1 camera being randomly dropped from every sample during training. We reason that sensor dropout forces the model to learn the correlation between images on different cameras, similar to other variants of dropout \cite{dropout} \cite{blockdrop}. We show on the left of Figure~\ref{fig:robust} that training the model with noisy extrinsics can lead to better test-time performance. For low amounts of noise at test-time, models that are trained without any noise in the extrinsics perform the best because the BEV CNN can trust the location of the splats with more confidence. For high amounts of extrinsic noise, our model sustains its good performance.

\begin{figure}[t] 
    \begin{center}
        \includegraphics[width=0.75\linewidth,trim=0 30 0 25,clip]{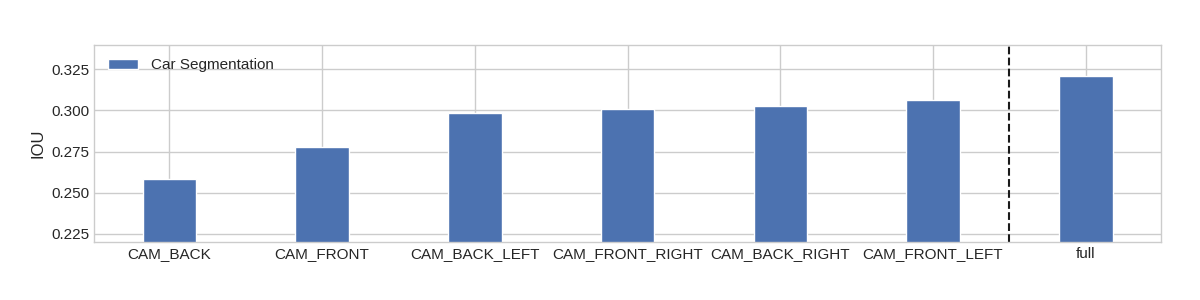}
    \end{center}
        \caption{\footnotesize We measure intersection-over-union of car segmentation when each of the cameras is missing. The backwards camera on the nuScenes camera rig has a wider field of view so it is intuitive that losing this camera causes the biggest decrease in performance relative to performance given the full camera rig (labeled ``full'' on the right). }
    \label{fig:drop}
    \end{figure}

\begin{figure}[t] 
        \begin{center}
            \includegraphics[width=0.95\linewidth]{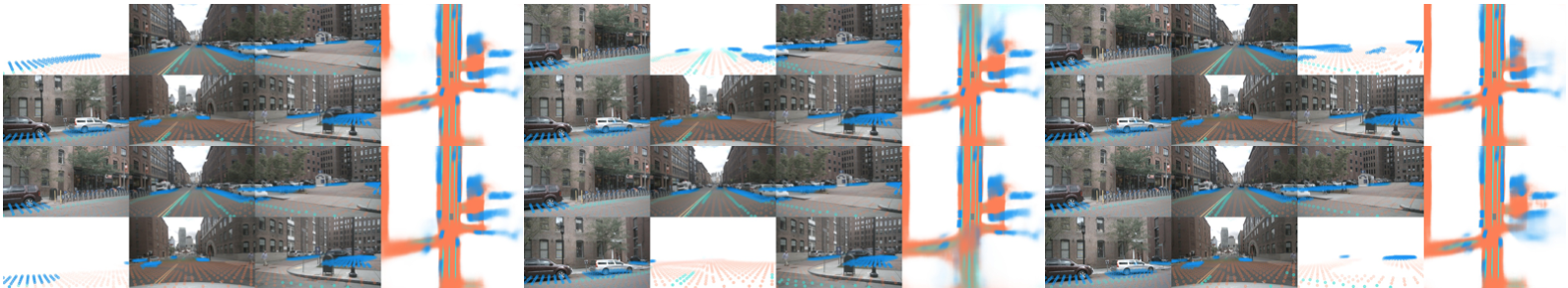} 
        \end{center}
            \caption{\footnotesize For a single time stamp, we remove each of the cameras and visualize how the loss the cameras effects the prediction of the network. Region covered by the missing camera becomes fuzzier in every case. When the front camera is removed (top middle), the network extrapolates the lane and drivable area in front of the ego and extrapolates the body of a car for which only a corner can be seen in the top right camera.}
        \label{fig:remove}
        \end{figure}

In Figure~\ref{fig:drop}, we measure the ``importance'' of each camera for the performance of car segmentation on nuScenes. Note that losing cameras on nuScenes implies that certain regions of the region local to the car have no sensor measurements and as a result performance strictly upper bounded by performance with the full sensor rig. Qualitative examples in which the network inpaints due to missing cameras are shown in Figure~\ref{fig:remove}. In this way, we measure the importance of each camera, suggesting where sensor redundancy is more important for safety.

\subsection{Zero-Shot Camera Rig Transfer}
\label{subsec:zeroshot}
 We now probe the generalization capabilities of Lift-Splat. In our first experiment, we measure performance of our model when only trained on images from a subset of cameras from the nuScenes camera rig but at test time has access to images from the remaining two cameras. In Table~\ref{tab:newcams}, we show that the performance of our model for car segmentation improves when additional cameras are available at test time without any retraining. %

 \begin{table}[t!]
 \begin{minipage}{0.23\linewidth}
    \begin{center}
      \begin{adjustbox}{width=0.9\textwidth}
    \begin{tabular}{|c|c|}
    \hline
        & IOU \\
    \hline
    4 & 26.53  \\
    \hline
    $4 + 1_{fl}$ & 27.35 \\
    \hline
    $4 + 1_{bl}$ & 27.27 \\
    \hline
    $4 + 1_{bl} + 1_{fl}$ & \textbf{27.94}\\
    \hline
    \end{tabular}
    \end{adjustbox}
    \end{center}
    \end{minipage}
     \begin{minipage}{0.745\linewidth}
    \caption{\footnotesize We train on images from only 4 of the 6 cameras in the nuScenes dataset. We then evaluate with the new cameras ($1_{bl}$ corresponds to the ``back left'' camera and $1_{fl}$ corresponds to the ``front left'' camera) and find that the performance of the model strictly increases as we add more sensors unseen during training.}
    \label{tab:newcams}
     \end{minipage}
    \end{table}

\begin{figure}[t] 
        \begin{center}
            \includegraphics[width=0.95\linewidth]{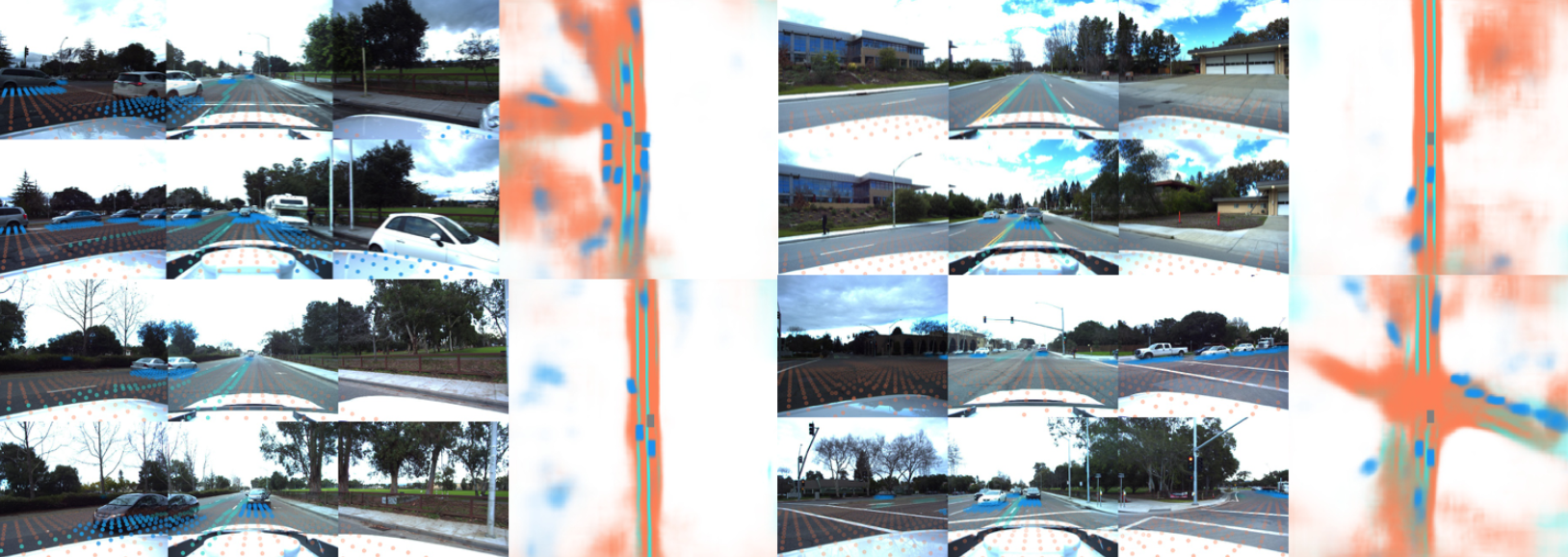}
        \end{center}
            \caption{\footnotesize We qualitatively show how our model performs given an entirely new camera rig at test time. Road segmentation is shown in orange, lane segmentation is shown in green, and vehicle segmentation is shown in blue.}
        \label{fig:zero}
        \end{figure}

 We take the above experiment a step farther and probe how well our model generalizes to the Lyft camera rig if it was only trained on nuScenes data. Qualitative results of the transfer are shown in Figure~\ref{fig:zero} and the benchmark against the generalization of our baselines is shown in Table~\ref{tab:zeroshot}.

 \begin{table}[t!]
      \begin{minipage}{0.58\linewidth}
\caption{\footnotesize We train the model on nuScenes then evaluate it on Lyft. The Lyft cameras are entirely different from the nuScenes cameras but the model succeeds in generalizing far better than the baselines. Note that our model has widened the gap from the standard benchmark in Tables~\ref{tab:obj} and~\ref{tab:map}.}
\label{tab:zeroshot}
     \end{minipage}
  \begin{minipage}{0.4\linewidth}
\begin{center}
   \begin{adjustbox}{width=\textwidth}
\begin{tabular}{|c|c|c|}
\hline
    & Lyft Car & Lyft Vehicle \\
\hline
CNN & 7.00 & 8.06 \\
\hline
Frozen Encoder & 15.08 & 15.82 \\
\hline
OFT & 16.25 & 16.27 \\
\hline
Lift-Splat (Us) & \textbf{21.35} & \textbf{22.59} \\
\hline
\end{tabular}
 \end{adjustbox}
\end{center}
     \end{minipage}
\end{table}

\subsection{Benchmarking Against Oracle Depth}
\label{subsec:oracle}
We benchmark our model against the pointpillars \cite{pointpillars} architecture which uses ground truth depth from LIDAR point clouds. As shown in Table~\ref{tab:maplidar}, across all tasks, our architecture performs slightly worse than pointpillars trained with a single scan of LIDAR. However, at least on drivable area segmentation, we note that we approach the performance of LIDAR. In the world in general, not all lanes are visible in a lidar scan. We would like to measure performance in a wider range of environments in the future.

\begin{figure}[t]%
        \centering
        \subfloat[IOU versus distance]{{\includegraphics[width=4cm]{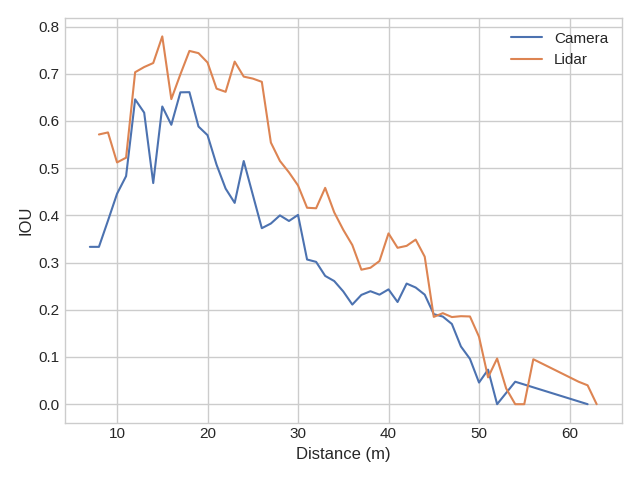} }}%
        \qquad
        \subfloat[IOU versus weather]{{\includegraphics[width=4cm]{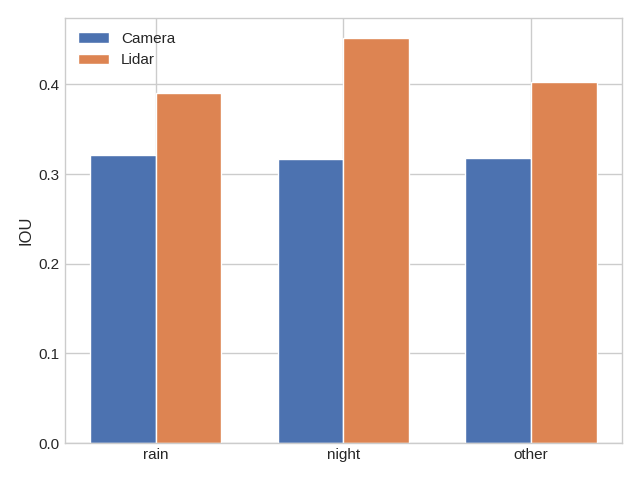} }}%
        \caption{\footnotesize We compare how our model's performance varies over depth and weather. As expected, our model drops in performance relative to pointpillars at nighttime.}%
        \label{fig:control}%
    \end{figure}

To gain insight into how our model differs from LIDAR, we plot how performance of car segmentation varies with two control variates: distance to the ego vehicle and weather conditions. We determine the weather of a scene from the description string that accompanies every scene token in the nuScenes dataset. The results are shown in Figure~\ref{fig:control}. We find that performance of our model is much worse than pointpillars on scenes that occur at night as expected. We also find that both models experience roughly linear performance decrease with increased depth.

\begin{table}[t!]
    \begin{center}
      \begin{adjustbox}{width=\textwidth}
      \addtolength{\tabcolsep}{3.5pt}
    \begin{tabular}{|l|c|c||c|c||c|c|}
    \cline{4-7}
    \multicolumn{3}{c|}{} & \multicolumn{2}{c||}{nuScenes} & \multicolumn{2}{c|}{Lyft}\\
    \cline{2-7}
    \multicolumn{1}{c|}{} & Drivable Area & Lane Boundary & Car & Vehicle & Car & Vehicle \\
    \hline
    Oracle Depth (1 scan) & 74.91 & 25.12 & 40.26 & 44.48 & 74.96 & 76.16 \\
    \hline
    Oracle Depth ($>1$ scan) & 76.96 & 26.80 & 45.36 & 49.51 & 75.42 & 76.49 \\
    \hline
    \hline
    Lift-Splat (Us) & 70.81 & 19.58 & 32.06 & 32.07 & 43.09 & 44.64 \\
    \hline
    \end{tabular}
    \end{adjustbox}
    \end{center}
    \caption{\footnotesize When compared to models that use oracle depth from lidar, there is still room for improvement. Video inference from camera rigs is likely necessary to acquire the depth estimates necessary to surpass lidar.}
    \label{tab:maplidar}
        \end{table}

\subsection{Motion Planning}
\label{subsec:planning}

Finally, we evaluate the capability of our model to perform planning by training the representation output by Lift-Splat to be a cost function. The trajectories that we generate are 5 seconds long spaced by 0.25 seconds. To acquire templates, we fit K-Means for $K=1000$ to all ego trajectories in the training set of nuScenes. At test time, we measure how well the network is able to predict the template that is closest to the ground truth trajectory under the L2 norm. This task is an important experiment for self-driving because the ground truth targets for this experiment are orders of magnitude less expensive to acquire than ground truth 3D bounding boxes. %
This task is also important for benchmarking the performance of camera-based approaches versus lidar-based approaches because although the ceiling for 3D object detection from camera-only is certainly upper bounded by lidar-only, the optimal planner using camera-only should in principle upper bound the performance of an optimal planner trained from lidar-only.

\begin{figure}[t] 
    \begin{center}
        \includegraphics[width=0.9\linewidth]{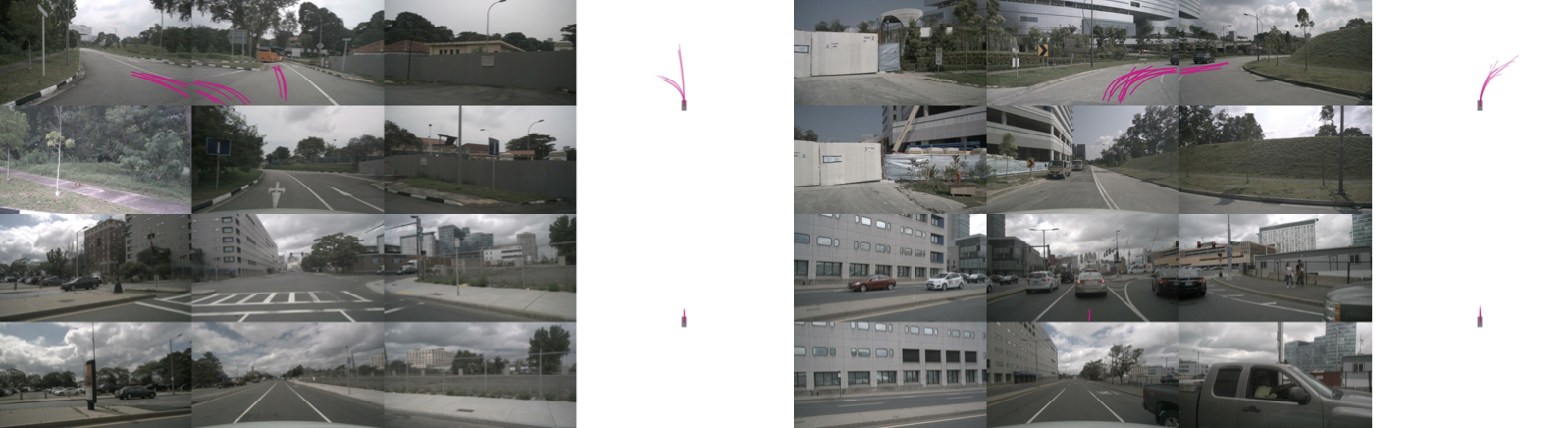}
    \end{center}
        \caption{\footnotesize We display the top 10 ranked trajectories out of the 1k templates. Video sequences are provided on our \href{https://nv-tlabs.github.io/lift-splat-shoot/}{project page}. Our model predicts bimodal distributions and curves from observations from a single timestamp. Our model does not have access to the speed of the car so it is compelling that the model predicts low-speed trajectories near crosswalks and brake lights.}
    \label{fig:qtraj}
    \end{figure}

Qualitative results of the planning experiment are shown in Figure~\ref{fig:qtraj}. The empirical results benchmarked against pointpillars are shown in Table~\ref{tab:planning}. The output trajectories exhibit desirable behavior such as following road boundaries and stopping at crosswalks or behind braking vehicles.

\begin{table}[t!]
\begin{minipage}{0.39\linewidth}
\begin{center}
  \begin{adjustbox}{width=\textwidth}
\begin{tabular}{|c|c|c|c|}
\hline
    & Top 5 & Top 10 & Top 20 \\
\hline
Lidar (1 scan) & 19.27 & 28.88 & 41.93 \\
\hline
Lidar (10 scans) & 24.99 & 35.39 & 49.84 \\
\hline
\hline
Lift-Splat (Us) & 15.52 & 19.94 & 27.99 \\
\hline
\end{tabular}
 \end{adjustbox}
\end{center}
\end{minipage}
\begin{minipage}{0.6\linewidth}
\caption{\footnotesize Since planning is framed as classification among a set of 1K template trajectories, we measure top-5, top-10, and top-20 accuracy. We find that our model is still lagging behind lidar-based approaches in generalization. Qualitative examples of the trajectories output by our model are shown in Fig.~\ref{fig:qtraj}.}
\label{tab:planning}
\end{minipage}
\end{table}

\section{Conclusion}
In this work, we present an architecture designed to infer bird's-eye-view representations from arbitrary camera rigs. Our model outperforms baselines on a suite of benchmark segmentation tasks designed to probe the model's ability to represent semantics in the bird's-eye-view frame without any access to ground truth depth data at training or test time. We present methods for training our model that make the network robust to simple models of calibration noise. Lastly, we show that the model enables end-to-end motion planning that follows the trajectory shooting paradigm. In order to meet and possibly surpass the performance of similar networks that exclusively use ground truth depth data from pointclouds, future work will need to condition on multiple time steps of images instead of a single time step as we consider in this work.

\bibliographystyle{splncs04}
\bibliography{egbib}

\begin{thebibliography}{10}
\providecommand{\url}[1]{\texttt{#1}}
\providecommand{\urlprefix}{URL }
\providecommand{\doi}[1]{https://doi.org/#1}

\bibitem{segnet}
Badrinarayanan, V., Kendall, A., Cipolla, R.: Segnet: {A} deep convolutional
  encoder-decoder architecture for image segmentation. CoRR
  \textbf{abs/1511.00561} (2015), \url{http://arxiv.org/abs/1511.00561}

\bibitem{nuscenes}
Caesar, H., Bankiti, V., Lang, A.H., Vora, S., Liong, V.E., Xu, Q., Krishnan,
  A., Pan, Y., Baldan, G., Beijbom, O.: nuscenes: {A} multimodal dataset for
  autonomous driving. CoRR  \textbf{abs/1903.11027} (2019),
  \url{http://arxiv.org/abs/1903.11027}

\bibitem{argo}
Chang, M.F., Ramanan, D., Hays, J., Lambert, J., Sangkloy, P., Singh, J., Bak,
  S., Hartnett, A., Wang, D., Carr, P., et~al.: Argoverse: 3d tracking and
  forecasting with rich maps. IEEE/CVF Conference on Computer Vision and
  Pattern Recognition (CVPR)  (Jun 2019)

\bibitem{mono3d}
Chen, X., Kundu, K., Zhang, Z., Ma, H., Fidler, S., Urtasun, R.: Monocular 3d
  object detection for autonomous driving. In: IEEE/CVF Conference on Computer
  Vision and Pattern Recognition (CVPR). pp. 2147--2156 (06 2016)

\bibitem{blockdrop}
Ghiasi, G., Lin, T., Le, Q.V.: Dropblock: {A} regularization method for
  convolutional networks. CoRR  \textbf{abs/1810.12890} (2018),
  \url{http://arxiv.org/abs/1810.12890}

\bibitem{goodfellow_text}
Goodfellow, I., Bengio, Y., Courville, A.: Deep Learning. MIT Press (2016),
  \url{http://www.deeplearningbook.org}

\bibitem{maskrcnn}
He, K., Gkioxari, G., Doll{\'{a}}r, P., Girshick, R.B.: Mask {R-CNN}. CoRR
  \textbf{abs/1703.06870} (2017), \url{http://arxiv.org/abs/1703.06870}

\bibitem{resnet}
He, K., Zhang, X., Ren, S., Sun, J.: Deep residual learning for image
  recognition. CoRR  \textbf{abs/1512.03385} (2015),
  \url{http://arxiv.org/abs/1512.03385}

\bibitem{zoox}
Hendy, N., Sloan, C., Tian, F., Duan, P., Charchut, N., Xie, Y., Wang, C.,
  Philbin, J.: Fishing net: Future inference of semantic heatmaps in grids
  (2020)

\bibitem{batchnorm}
Ioffe, S., Szegedy, C.: Batch normalization: Accelerating deep network training
  by reducing internal covariate shift. CoRR  \textbf{abs/1502.03167} (2015),
  \url{http://arxiv.org/abs/1502.03167}

\bibitem{trans_invariance}
Kayhan, O.S., Gemert, J.C.v.: On translation invariance in cnns: Convolutional
  layers can exploit absolute spatial location. In: Proceedings of the IEEE/CVF
  Conference on Computer Vision and Pattern Recognition (CVPR) (June 2020)

\bibitem{ssd16}
Kehl, W., Manhardt, F., Tombari, F., Ilic, S., Navab, N.: {SSD-6D:} making
  rgb-based 3d detection and 6d pose estimation great again. CoRR
  \textbf{abs/1711.10006} (2017)

\bibitem{lyftlevel5}
Kesten, R., Usman, M., Houston, J., Pandya, T., Nadhamuni, K., Ferreira, A.,
  Yuan, M., Low, B., Jain, A., Ondruska, P., Omari, S., Shah, S., Kulkarni, A.,
  Kazakova, A., Tao, C., Platinsky, L., Jiang, W., Shet, V.: Lyft level 5 av
  dataset 2019. url{https://level5.lyft.com/dataset/} (2019)

\bibitem{adam}
Kingma, D.P., Ba, J.: Adam: A method for stochastic optimization. CoRR
  \textbf{abs/1412.6980} (2014)

\bibitem{panoptic}
Kirillov, A., He, K., Girshick, R.B., Rother, C., Doll{\'{a}}r, P.: Panoptic
  segmentation. CoRR  \textbf{abs/1801.00868} (2018),
  \url{http://arxiv.org/abs/1801.00868}

\bibitem{cifar}
Krizhevsky, A.: Learning multiple layers of features from tiny images (2009)

\bibitem{alexnet}
Krizhevsky, A., Sutskever, I., Hinton, G.E.: Imagenet classification with deep
  convolutional neural networks. In: Pereira, F., Burges, C.J.C., Bottou, L.,
  Weinberger, K.Q. (eds.) Advances in Neural Information Processing Systems 25,
  pp. 1097--1105. Curran Associates, Inc. (2012),
  \url{http://papers.nips.cc/paper/4824-imagenet-classification-with-deep-convolutional-neural-networks.pdf}

\bibitem{pointpillars}
Lang, A.H., Vora, S., Caesar, H., Zhou, L., Yang, J., Beijbom, O.:
  Pointpillars: Fast encoders for object detection from point clouds. CoRR
  \textbf{abs/1812.05784} (2018)

\bibitem{mnist}
Lecun, Y., Bottou, L., Bengio, Y., Haffner, P.: Gradient-based learning applied
  to document recognition. In: Proceedings of the IEEE. pp. 2278--2324 (1998)

\bibitem{neuralvolume}
Lombardi, S., Simon, T., Saragih, J., Schwartz, G., Lehrmann, A., Sheikh, Y.:
  Neural volumes. ACM Transactions on Graphics  \textbf{38}(4),  1–14 (Jul
  2019). \doi{10.1145/3306346.3323020},
  \url{http://dx.doi.org/10.1145/3306346.3323020}

\bibitem{krishna}
Mani, K., Daga, S., Garg, S., Shankar, N.S., Jatavallabhula, K.M., Krishna,
  K.M.: Monolayout: Amodal scene layout from a single image. ArXiv
  \textbf{abs/2002.08394} (2020)

\bibitem{relu}
Nair, V., Hinton, G.E.: Rectified linear units improve restricted boltzmann
  machines. In: ICML (2010)

\bibitem{pytorch}
Paszke, A., Gross, S., Massa, F., Lerer, A., Bradbury, J., Chanan, G., Killeen,
  T., Lin, Z., Gimelshein, N., Antiga, L., Desmaison, A., K{\"o}pf, A., Yang,
  E., DeVito, Z., Raison, M., Tejani, A., Chilamkurthy, S., Steiner, B., Fang,
  L., Bai, J., Chintala, S.: Pytorch: An imperative style, high-performance
  deep learning library. In: NeurIPS (2019)

\bibitem{fastdraw}
Philion, J.: Fastdraw: Addressing the long tail of lane detection by adapting a
  sequential prediction network. In: Proceedings of the IEEE/CVF Conference on
  Computer Vision and Pattern Recognition (CVPR) (June 2019)

\bibitem{pkl}
Philion, J., Kar, A., Fidler, S.: Learning to evaluate perception models using
  planner-centric metrics. In: Proceedings of the IEEE/CVF Conference on
  Computer Vision and Pattern Recognition (CVPR) (June 2020)

\bibitem{fastsingle}
Poirson, P., Ammirato, P., Fu, C., Liu, W., Kosecka, J., Berg, A.C.: Fast
  single shot detection and pose estimation. CoRR  \textbf{abs/1609.05590}
  (2016)

\bibitem{monogr}
Qin, Z., Wang, J., Lu, Y.: Monogrnet: A geometric reasoning network for
  monocular 3d object localization. Proceedings of the AAAI Conference on
  Artificial Intelligence  \textbf{33},  8851--8858 (07 2019).
  \doi{10.1609/aaai.v33i01.33018851}

\bibitem{roddick}
Roddick, T., Cipolla, R.: Predicting semantic map representations from images
  using pyramid occupancy networks. In: Proceedings of the IEEE/CVF Conference
  on Computer Vision and Pattern Recognition (CVPR) (June 2020)

\bibitem{oft}
Roddick, T., Kendall, A., Cipolla, R.: Orthographic feature transform for
  monocular 3d object detection. CoRR  \textbf{abs/1811.08188} (2018)

\bibitem{imagenet}
Russakovsky, O., Deng, J., Su, H., Krause, J., Satheesh, S., Ma, S., Huang, Z.,
  Karpathy, A., Khosla, A., Bernstein, M., Berg, A.C., Fei-Fei, L.: Imagenet
  large scale visual recognition challenge (2014)

\bibitem{mair}
Simonelli, A., Bul{\`{o}}, S.R., Porzi, L., L{\'{o}}pez{-}Antequera, M.,
  Kontschieder, P.: Disentangling monocular 3d object detection. CoRR
  \textbf{abs/1905.12365} (2019)

\bibitem{lighthouse}
Srinivasan, P.P., Mildenhall, B., Tancik, M., Barron, J.T., Tucker, R.,
  Snavely, N.: Lighthouse: Predicting lighting volumes for spatially-coherent
  illumination (2020)

\bibitem{dropout}
Srivastava, N., Hinton, G., Krizhevsky, A., Sutskever, I., Salakhutdinov, R.:
  Dropout: A simple way to prevent neural networks from overfitting. Journal of
  Machine Learning Research  \textbf{15},  1929--1958 (2014)

\bibitem{splatnet}
Su, H., Jampani, V., Sun, D., Maji, S., Kalogerakis, E., Yang, M., Kautz, J.:
  Splatnet: Sparse lattice networks for point cloud processing. CoRR
  \textbf{abs/1802.08275} (2018), \url{http://arxiv.org/abs/1802.08275}

\bibitem{waymo}
Sun, P., Kretzschmar, H., Dotiwalla, X., Chouard, A., Patnaik, V., Tsui, P.,
  Guo, J., Zhou, Y., Chai, Y., Caine, B., Vasudevan, V., Han, W., Ngiam, J.,
  Zhao, H., Timofeev, A., Ettinger, S., Krivokon, M., Gao, A., Joshi, A.,
  Zhang, Y., Shlens, J., Chen, Z., Anguelov, D.: Scalability in perception for
  autonomous driving: Waymo open dataset (2019)

\bibitem{scnn}
Takikawa, T., Acuna, D., Jampani, V., Fidler, S.: Gated-scnn: Gated shape cnns
  for semantic segmentation. In: Proceedings of the IEEE/CVF International
  Conference on Computer Vision (ICCV) (October 2019)

\bibitem{efficientnet}
Tan, M., Le, Q.V.: Efficientnet: Rethinking model scaling for convolutional
  neural networks. CoRR  \textbf{abs/1905.11946} (2019),
  \url{http://arxiv.org/abs/1905.11946}

\bibitem{tucker2020singleview}
Tucker, R., Snavely, N.: Single-view view synthesis with multiplane images
  (2020)

\bibitem{pseudolidar}
Wang, Y., Chao, W., Garg, D., Hariharan, B., Campbell, M., Weinberger, K.Q.:
  Pseudo-lidar from visual depth estimation: Bridging the gap in 3d object
  detection for autonomous driving. CoRR  \textbf{abs/1812.07179} (2018)

\bibitem{pseudolidarpp}
You, Y., Wang, Y., Chao, W., Garg, D., Pleiss, G., Hariharan, B., Campbell, M.,
  Weinberger, K.Q.: Pseudo-lidar++: Accurate depth for 3d object detection in
  autonomous driving. CoRR  \textbf{abs/1906.06310} (2019)

\bibitem{nmp}
Zeng, W., Luo, W., Suo, S., Sadat, A., Yang, B., Casas, S., Urtasun, R.:
  End-to-end interpretable neural motion planner. IEEE/CVF Conference on
  Computer Vision and Pattern Recognition (CVPR) pp. 8652--8661 (2019)

\end{thebibliography}
\end{document}